\documentclass[letterpaper]{article} 
\usepackage{aaai25}  
\usepackage{times}  
\usepackage{helvet}  
\usepackage{courier}  
\usepackage[hyphens]{url}  
\usepackage{graphicx} 
\usepackage{natbib}  
\usepackage{caption} 
\usepackage{algorithm}
\usepackage{algorithmic}

\usepackage{amsmath}
\usepackage[capitalize,noabbrev]{cleveref}
\usepackage{amssymb}

\usepackage{booktabs}
\usepackage{multirow}
\usepackage{rotating}   
\usepackage{booktabs}
\usepackage{colortbl}
\usepackage{graphicx}
\usepackage[table]{xcolor}
\usepackage{xspace}
\newcommand{\papertitle}{\texttt{CodeGraph}\xspace}

\definecolor{grey}{rgb}{0.5, 0.5, 0.5}
\definecolor{maroon}{rgb}{0.5, 0.0, 0.0}

\newcommand{\method}[1]{\acr{#1}}
\newcommand{\acr}[1]{{{\textsc{#1}}}}

\usepackage[most]{tcolorbox} 
\usepackage{listings}
\renewcommand\fbox{\fcolorbox{blue!50}{blue!10}}

\NewDocumentCommand{\shizhe}
{ mO{} }{\textcolor{blue}{\textsuperscript{\textit{shizhe}}\textsf{\textbf{\small[#1]}}}}

\usepackage{newfloat}
\usepackage{listings}
\DeclareCaptionStyle{ruled}{labelfont=normalfont,labelsep=colon,strut=off} 
\floatstyle{ruled}
\newfloat{listing}{tb}{lst}{}
\floatname{listing}{Listing}
\title{\papertitle: Enhancing Graph Reasoning of LLMs with Code}
\author{
    Qiaolong Cai\equalcontrib, Zhaowei Wang\equalcontrib, Shizhe Diao, James Kwok, Yangqiu Song \\
}
\affiliations{
    Department of Computer Science and Engineering, HKUST, Hong Kong SAR, China\\
    \{qcaiaj, zwanggy, sdiaoaa, jamesk, yqsong\}@cse.ust.hk

}

\usepackage{bibentry}

\begin{document}

\maketitle


\begin{abstract}
With the increasing popularity of large language models (LLMs), reasoning on basic graph algorithm problems is an essential intermediate step in assessing their abilities to process and infer complex graph reasoning tasks. Existing methods usually convert graph-structured data to textual descriptions and then use LLMs for reasoning and computation. However, LLMs often produce computation errors on arithmetic parts in basic graph algorithm problems, such as counting the number of edges. In addition, they struggle to control or understand the output of the reasoning process, raising concerns about whether LLMs are simply guessing. In this paper, we introduce \papertitle, a method that encodes graph problem solutions as code. The methods solve new graph problems by learning from exemplars, generating programs, and executing them via a program interpreter. Using the few-shot setting, we evaluate \papertitle with the base LLM being GPT-3.5 Turbo, Llama3-70B Instruct, Mixtral-8x22B Instruct, and Mixtral-8x7B Instruct.
Experimental results on six tasks with six graph encoding methods in the GraphQA dataset demonstrate that \papertitle can boost performance on graph reasoning tasks inside LLMs by 1.3\% to 58.6\%, depending on the task. Compared to the existing methods, \papertitle demonstrates strong performance on arithmetic problems in graph tasks and offers a more controllable and interpretable approach to the reasoning process.

\end{abstract}

\begin{figure*}[t]
    \centering
    \includegraphics[width=0.9\textwidth]{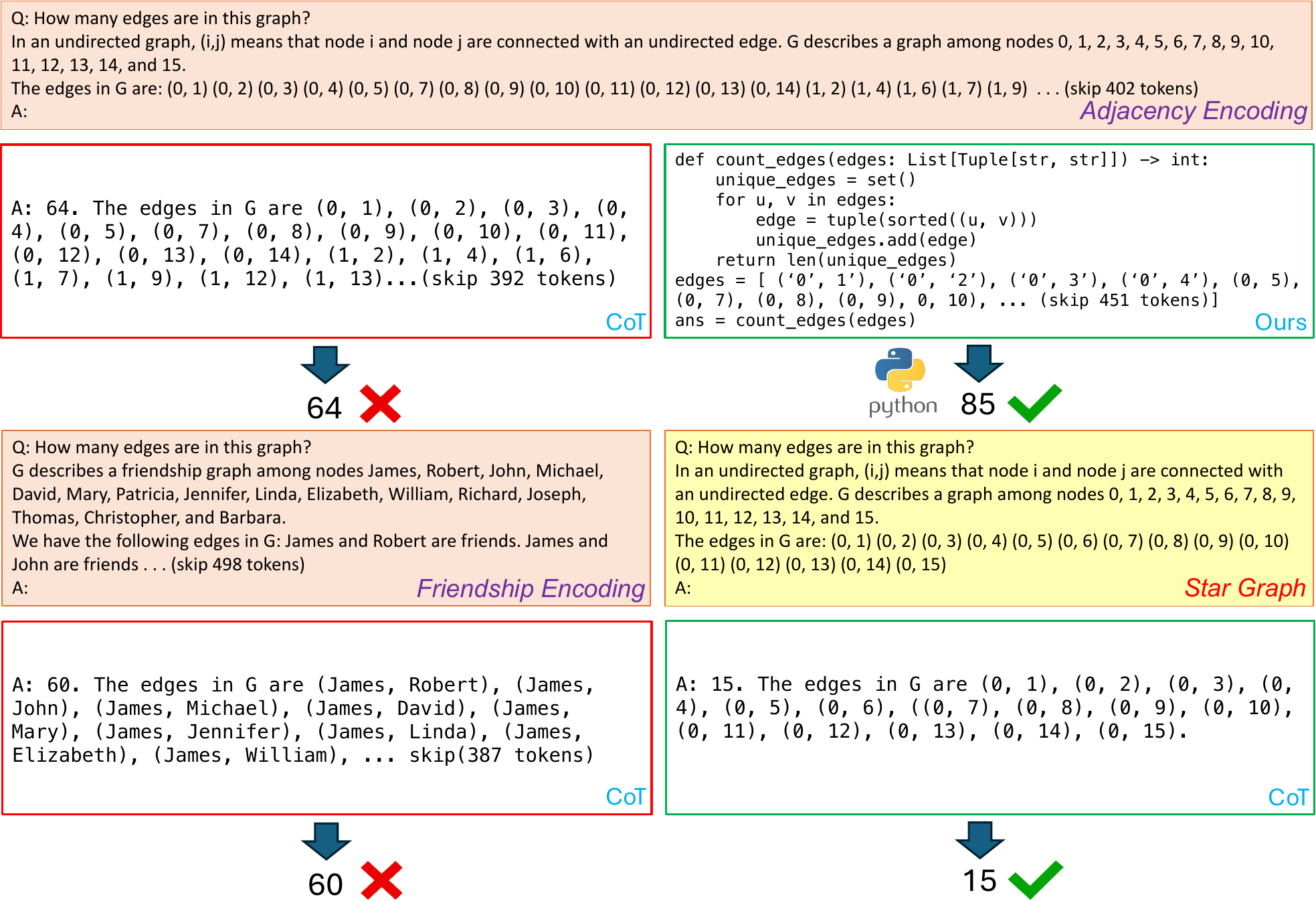}
    \caption{Comparison between CoT and our method inspired from PAL~\cite{gao2023pal} and PoT~\cite{chen2023program}.
    In the upper example, the limitation of LLMs in handling arithmetic within the graph task is highlighted, while our method addresses this issue through code and external execution. 
    In the lower-left example, LLM performance is shown to be sensitive to prompt templates that convert graph structures into natural language.
    The lower-right example demonstrates that the LLM’s performance is significantly influenced by the graph structure.
    }
    \label{fig:cot_pot}
\end{figure*}

%

\section{Introduction}

The natural language processing (NLP) community has witnessed the recent advancement of large language models (LLMs)~\cite{radford_language_2019, Brown2020LanguageMA, Ouyang2022TrainingLM, Touvron2023LLaMAOA, Wei2021FinetunedLM}, which achieve state-of-the-art results~\cite{mao2023gptevalsurveyassessmentschatgpt} in diverse tasks such as code generation~\cite{bubeck2023sparksartificialgeneralintelligence}, information retrieval~\cite{sun2023chatgptgoodsearchinvestigating}, and question answering~\cite{gemmateam2024gemmaopenmodelsbased}. 
However, sole reliance on natural language sometimes leads to hallucination~\cite{huang2023survey}, where results deviate from user inputs. 
A common solution is to adapt external information after training dates to answer questions that require dynamically changing knowledge~\cite{vu2023freshllmsrefreshinglargelanguage}. 
Among external information, graph-structured data can flexibly represent additional factual and fresh data, which is one of the most flexible ways to mitigate the hallucination issue~\cite{lewis2021retrievalaugmentedgenerationknowledgeintensivenlp,guu2020retrieval,Pan_2024}. 
LLMs have been shown to adapt their parametric knowledge when supplied with new graphs~\cite{kadavath2022languagemodelsmostlyknow}.
Thus, an underlying yet crucial question attracts more and more research attention~\cite{wang2024languagemodelssolvegraph}:
{\em Can LLMs reason with graphs?}

To answer this question, most studies map graphs to textual description and adapt natural language with prompting heuristics, such as Chain of Thought (CoT)~\cite{wei2023chainofthought}, to solve graph algorithm problems using generative LLMs~\cite{ye2024language, liu2023evaluating,huang2024llms, tang2024graphgpt}.
Some studies also extend the input texts with soft tokens~\cite{perozzi2024letgraphtalkingencoding}. Those textual description-based approaches convert node and edge lists in a graph into plain text with prompts~\cite{liu2023evaluatinglargelanguagemodels, zhang2024llm4dyglargelanguagemodels,wang2024languagemodelssolvegraph}. Among them, \citet{fatemi2023talk} further explores various prompts (such as representing a node as a name, with edges indicating various relationships like friendships or co-authorships) to express the graph structure in natural language. However, those approaches utilize LLMs for both reasoning and computation in natural language, i.e., the language model not only needs to generate the texts for constructing graphs using nodes and edges but also needs to perform computation for some basic graph tasks such as \textit{node degree} and \textit{edge count}. We argue that text descriptions are not ideal for solving basic graph tasks based on the findings in our experiments: (1) LLMs are prone to arithmetic calculation errors, such as counting the degree of a node and counting the number of edges. (2) LLMs' performance on basic graph tasks is highly sensitive to the prompt template that converts graphs to natural language. (3) Different graph structures significantly influence the effectiveness of LLMs when assessing graph-related tasks. For instance, on the 
\textit{cycle check} task, LLMs tend to have a strong bias toward graphs containing cycles. This bias can lead to lower accuracies when analyzing acyclic structures such as path graphs and higher accuracies with complete graphs, which always contain cycles.

In this work, we propose \papertitle, a prompting framework that generates solutions in code using an LLM for basic graph tasks and delegates the computation steps to an external language interpreter. In \papertitle, LLMs articulate the reasoning process as Python programs. A Python interpreter is then employed to accomplish the computation and provide the final answer. We illustrate the difference between CoT methods in previous works and \papertitle in \cref{fig:cot_pot}. In the upper example, there are 85 edges in the graph, and CoT leads to extremely different answers. In the lower-left example, the prompt describes the same graph as in the upper example but with a different text representation, which affects the CoT result. In the lower-right example, the graph uses the same encoding function as in the upper example but with a simpler structure, which allows the CoT to deduce the correct answer successfully.

We conduct comprehensive studies to examine the effectiveness of the proposed \papertitle\footnote{The code is available at \url{https://github.com/HKUST-KnowComp/CodeGraph}.} in basic graph tasks using the GraphQA benchmark~\cite{fatemi2023talk}. The first study evaluates the performance of code across six basic graph tasks 
(such as \textit{edge existence} and \textit{node degree} tasks)
with various graph encoding functions.
The graphs used here are generated by the ER model~\cite{ErdosRenyi1959}. The proposed method significantly increases the average accuracy on the six tasks from 63.3\% (zero-shot) to 96.1\% via GPT-3.5 Turbo with one-shot learning. Then, we show the robustness of 
\papertitle
across six additional graph structures 
besides ER. Further, we show that 
\papertitle
can work with LLMs other than GPT-3.5 Turbo, such as Llama3-70B Instruct, Mixtral-8x22B Instruct, and Mixtral-8x7B Instruct. The smallest model uses only 14 billion activated parameters. Lastly, we illustrate that by providing more exemplars,
\papertitle can achieve higher performance of smaller models with weaker coding capabilities.

\section{Related Work}
Recently, research on combining graph learning with large language models has become a rapidly growing area. The community has dedicated various works to studying multiple tasks, including node classification~\cite{chen2024exploring, ye2024language}, reasoning in KG~\cite{do2024constraintchecker}, KG construction~\cite{neuhaus2023ontologies}, molecular learning~\cite{le2024molx}, and others~\cite{li2023multi}. Among them, our work focuses on utilizing LLMs' coding ability to solve basic graph tasks with the following relevant fields:

\paragraph{Basic Graph Tasks with LLMs:} \citet{fatemi2023talk} introduced a graph benchmark to evaluate LLMs' performance on basic graph tasks, including \textit{edge existence}, \textit{node degree}, and \textit{cycle check}. Furthermore, \citet{wang2024language} introduced a larger dataset with more diverse and complex tasks, such as maximum flow and shortest path. While a lot of works try to enhance LLMs' reasoning ability on graphs~\cite{tang2023graphgpt, tang2024higpt, wei2024llmrec}, only a few methods are proposed on these benchmarks to enhance the LLMs' reasoning ability for basic graph tasks. \citet{perozzi2024letgraphtalkingencoding} proposed to learn an encoding function to generate fine-tuned soft-tokens to encode structural information of graphs. \citet{chen2024graphwiz} proposed a dataset of basic graph tasks to instruct tune LLMs for better performance. In contrast to these works, this paper seeks to solve basic graph tasks using the coding ability of LLMs and does not require high computation costs to fine-tune LLMs.

\paragraph{Code-based reasoning with LLMs:}
Today's prominent LLMs
differ from past language models not
only in size but also in that they are
trained on a combination of natural language
and formal language (code)~\cite{yang2024llmwizardcodewand}. The ability to generate code 
helps
LLMs to
improve their performance on various tasks, including commonsense graph understading~\cite{madaan2022languagemodelscodefewshot, wang2023code4structcodegenerationfewshot,chen2023vistructvisualstructuralknowledge}, tool using~\cite{schick2023toolformerlanguagemodelsteach, tang2023toolalpacageneralizedtoollearning,
patil2023gorillalargelanguagemodel, song2023restgptconnectinglargelanguage, hao2024toolkengptaugmentingfrozenlanguage}, task decomposition with Chain of Thought~\cite{fu2022gptobtaincotwhy,ma2023trainingstagedoescode}, and others~\cite{liang2023codepolicieslanguagemodel,vemprala2023chatgptroboticsdesignprinciples,ma2024lampilotopenbenchmarkdataset}.

Many 
of these works
try to improve LLMs' mathematical problem-solving and theorem-proving abilities \cite{ye2023largelanguagemodelsversatile,polu2020generativelanguagemodelingautomated,Drori_2022,cheng2023bindinglanguagemodelssymbolic,lei2024hintthoughtpromptingexplainable}. \citet{gao2023pal}, \citet{ye2023largelanguagemodelsversatile}, and \citet{chen2023program} introduced program-aided language models to evaluate LLMs' performance in arithmetic and symbolic reasoning tasks. They directly execute the code generated by LLMs, ensuring that the reasoning process adheres to the logic and mathematical rules, translating problem-solving steps into executable programs. Inspired by them, we propose \papertitle, which generates program language to solve basic graph tasks and delegates computation steps to an external language interpreter.

\begin{figure}[t]
    \centering
    \includegraphics[scale=0.324]{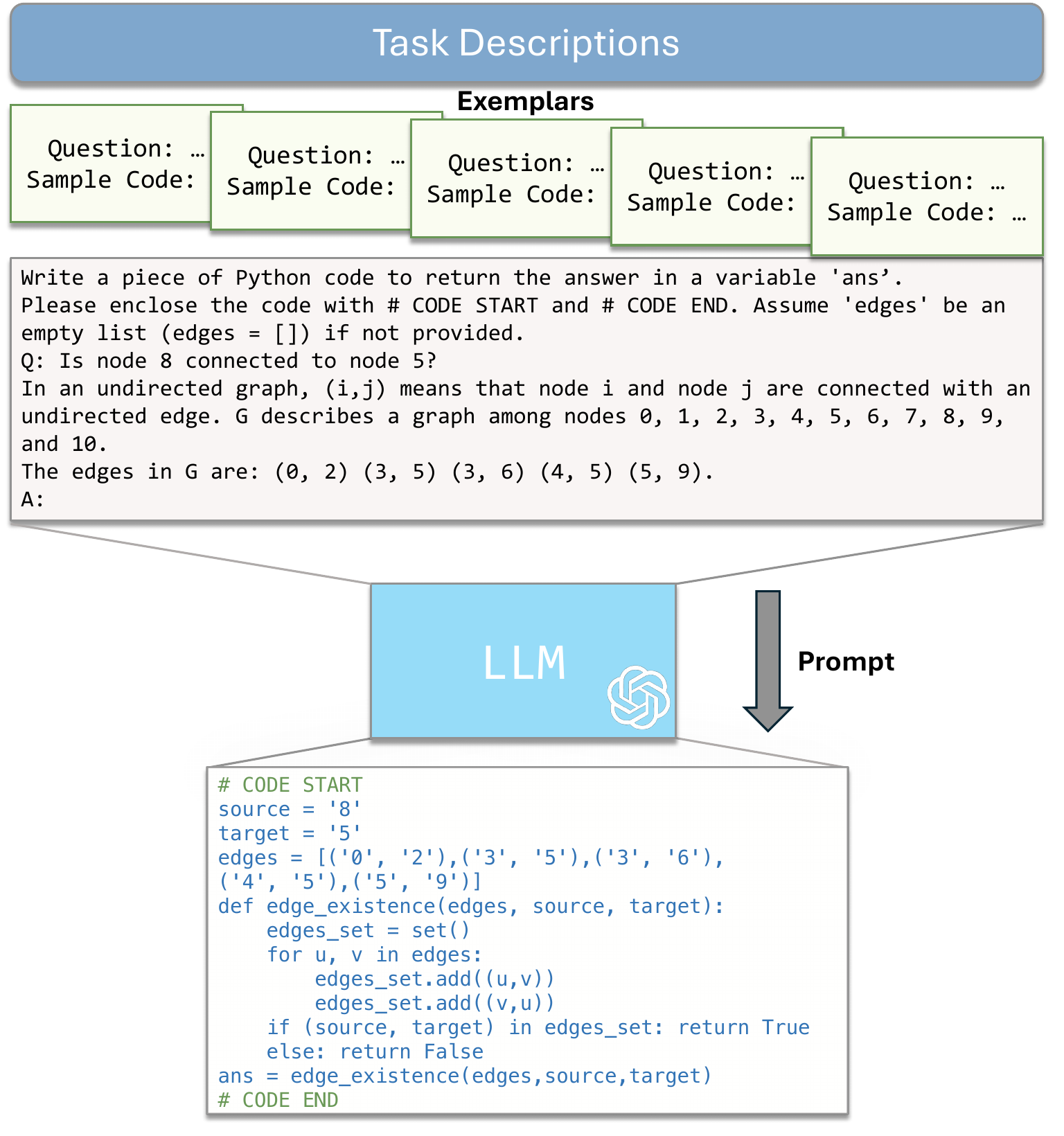}
    \caption{\papertitle in few-shot setting. We provide the LLM with a few examples and then ask it to convert the graph into Python code using the provided nodes and/or edges. Then, a Python function is generated to obtain the answer.}
    \label{fig:code_graph}
\end{figure}

\section{Task Formulation}
In this paper, we study the performance of LLMs on basic graph problems, including \textit{connected nodes}, \textit{cycle check}, \textit{edge existence}, \textit{edge count}, \textit{node degree}, and \textit{node count} (the definitions are in Appendix \ref{appendix:task_description}). 
Here, we treat the basic graph problems in a sequence-to-sequence format.

First, 
a graph 
$G = (V, E)$ 
and a basic graph problem $Q$
are provided as
input to the
LLM, where
$V$ 
denotes the vertices (nodes) and 
\( E \subseteq (V \times V) \) 
denotes the edges.
Each graph $G$ is 
then
converted to natural language using an encoding function by listing the nodes and edges in the graph. In the following, we use $f$ to denote an LLM, and
$A$ be
the 
answer 
(output)
for a given question $Q$, i.e., $A = f(G,Q).$




\section{Methodology}

In this section, we introduce the components of \papertitle, including its graph encoding functions, task descriptions, exemplars, and a test example.

\subsection{Graph Encoding Functions}
Following GraphQA~\cite{fatemi2023talk}, 
we translate graph-structured data into natural language using six graph encoding functions: adjacency, friendship, co-authorship, incident, social network, and expert encoding. These functions encode nodes as integers, names, or letters, and edges as relationships such as `friends' or `coworkers', forming the textual descriptions for questions in basic graph tasks. Below, we present the details and examples of the Adjacency and Friendship encoding functions. Descriptions of the other encoding functions can be found in Appendix~\ref{appendix:graph_encoding_function}.

\begin{itemize}
\label{encodingfunction:adjacency}
\item \textbf{Adjacency}: Nodes are encoded as integers, and edges are represented using parentheses.
\end{itemize}
\noindent\begin{center}
\noindent\fbox{%
\parbox{\dimexpr\columnwidth-0.5cm\relax}{
\textbf{Adjacency:}\\
In an undirected graph, (i,j) means that node i and node j are connected with an undirected edge.\\
G describes a graph among nodes 0, 1, 2, 3, and 4.\\
The edges in G are: (0, 1) (0, 2) (1, 2) (2, 3) (2, 4).
}
}
\end{center}

\begin{itemize}
\item \textbf{Friendship}: Nodes are represented by common English first names, with edges denoting friendships.
\end{itemize}
\noindent\fbox{%
\parbox{\dimexpr\columnwidth-0.5cm\relax}{
\textbf{Friendship:}\\ 
G describes a friendship graph among James, Robert, John, Michael, and David.\\
We have the following edges in G:\\
James and Robert are friends.\\
James and John are friends.\\
Robert and John are friends.\\
John and Michael are friends.\\
John and David are friends.
}
}

\subsection{\papertitle Method}

Figure \ref{fig:code_graph} 
shows the components of the proposed \papertitle, 
including task descriptions, exemplars, and a test example.
In the few-shot setting, a few exemplars of (question, sample code) pairs are prefixed as demonstrations to guide the LLM in generating a program depending on the graph task.

The task description 
begins by defining the role of the LLM, instructing it to generate code that solves a basic graph task using the given example. The task description then specifies the graph task to be solved and standardizes the expected input and output formats for the Python code (e.g., \textit{``Write a piece of Python code to return the answer in a variable \texttt{ans}. Please enclose the code with \texttt{\# CODE START} and \texttt{\# CODE END}. Assume \texttt{edges} to be an empty list (\texttt{edges = []}) if not provided."}) A complete example of the task description can be seen in Appendix~\ref{appendix:task_description}.


Exemplars follow the task description in the prompt template. As shown in Figure \ref{fig:code_graph}, each exemplar has two parts: a question and a sample code. The question begins with `Q:' and presents a graph question based on the task (e.g., \textit{``Is node 8 connected to node 5?''} for the \textit{edge existence} task). Then, the question describes a graph where nodes and edges are translated into natural language using encoding functions. The concrete prompt for graph questions is provided in Appendix~\ref{appendix:graph_questions}.

For each example, the sample code starts by mapping the text of the nodes and edges into code, which is stored in Python lists or adjacency lists, depending on the task. It then continues with `A:' and a Python function. Each graph task has a specific Python function, with nodes and/or edges as input parameters, depending on the task. 
An example function for the \textit{edge existence} task is shown at the bottom of Figure~\ref{fig:code_graph}. To ensure executions of the Python interpreter, we employ special tokens `\# CODE START' and `\# CODE END' to mark obvious boundaries of the Python function and its inputs. We propose different Python functions to solve the six basic graph tasks. The detailed sample code for each task is provided in Appendix~\ref{appendix:sample_programs}.

The final part of the LLM input is the test example. It follows the same format as the question in the exemplar but uses different graphs. Typically, it employs the same graph encoding function as the exemplar unless specified otherwise. It ends with `A:' to prompt the LLM to convert the graph into lists or adjacency lists, and then generate a program based on the converted graph.

\begin{table*}[t]
\centering
 \small
\setlength{\tabcolsep}{3pt}
\begin{tabular}{c|c|cccccc}
\toprule
\textbf{Method} & \textbf{Encoding} & \textbf{Edge Exist.} & \textbf{Node Deg.} & \textbf{Node Count} & \textbf{Edge Count} & \textbf{Conn. Nodes} & \textbf{Cycle Check} \\ 

\midrule
\multirow{7}{*}{\begin{sideways}\method{\normalsize Zero-shot}\end{sideways}}
&  \cellcolor{gray!25} Overall ($\mu / \delta$) &\cellcolor{gray!25} 72.9 / 8.2 &\cellcolor{gray!25} 49.5/31.0 &\cellcolor{gray!25} 97.4 / 6.60 &\cellcolor{gray!25} 34.1 / 40.8 &\cellcolor{gray!25} 40.6 / 50.2 &\cellcolor{gray!25} 85.3 / 3.80\\
& Adjacency & 75.4 & 49.2 & 100.0 & 44.8 & 42.6 & 84.8\\
& Incident & 73.0 & 73.0 &  100.0 & 4.00 & 78.0 & 83.6\\
& Co-authorship & 69.8 & 42.0 & 99.4 & 36.2 & 27.8 & 83.8\\
& Friendship & 69.0 & 45.6 & 93.4 & 41.0 & 31.8 & 87.4\\
& Social network & 73.2 & 43.6 & 93.4 & 41.8 & 29.8 & 85.8\\
& Expert & 77.2 & 43.4 & 98.0 & 37.0 & 33.8 & 86.6\\

\midrule
\multirow{7}{*}{\begin{sideways}\method{\normalsize One-shot}\end{sideways}} 
& \cellcolor{gray!25} Overall ($\mu / \delta$)&\cellcolor{gray!25} 71.8 / 5.0&\cellcolor{gray!25} 44.6 / 43.4&\cellcolor{gray!25} 85.6 / 28.6&\cellcolor{gray!25} 28.3 / 33.6&\cellcolor{gray!25} 36.5 / 36.0&\cellcolor{gray!25} 86.9 / 7.4\\
& Adjacency & 73.8 & 43.0 & 99.2 & 40.6 & 42.8 & 89.2\\
& Incident & 71.8 & 77.6 & 99.2 & 7.0 & 62.6 & 82.0\\
& Co-authorship & 68.8 & 34.2 & 76.8 & 23.0 & 26.6 & 87.0 \\
& Friendship & 70.0 & 34.8 & 70.6 & 35.8 & 28.0 & 84.4\\
& Social network & 73.0 & 35.6 & 73.6 & 36.4 & 29.4 & 89.4\\
& Expert & 73.2 & 42.6 & 94.0 & 27.0 & 29.4 & 89.4\\
\midrule
\multirow{7}{*}{\begin{sideways}\method{\normalsize CoT (1-shot)}\end{sideways}}
& \cellcolor{gray!25} Overall ($\mu / \delta$)&\cellcolor{gray!25} 72.5 / 4.2&\cellcolor{gray!25} 42.8 / 50.6&\cellcolor{gray!25} 81.2 / 37.4&\cellcolor{gray!25} 26.1 / 30.0&\cellcolor{gray!25} 37.2 / 37.2&\cellcolor{gray!25} 82.6 / 10.6\\
& Adjacency & 72.8 & 44.8 & 99.6 & 40.2 & 40.4 & 86.8 \\
& Incident & 74.4 & 79.6 & 99.6 & 10.2 & 63.6 & 76.6 \\
& Co-authorship & 70.2 & 29.0 & 62.2 & 21.0 & 26.4 & 81.8 \\
& Friendship & 71.8 & 31.2 & 63.0 & 29.8 & 30.2 & 81.2\\
& Social network & 72.6 & 32.4 & 64.8 & 30.2 & 30.2 & 87.2 \\
& Expert & 73.0 & 40.0 & 98.0 & 25.2 & 32.6 & 81.8 \\
\midrule
\multirow{7}{*}{\begin{sideways}\method{\normalsize Ours(1-shot)}\end{sideways}}
& \cellcolor{gray!25} Overall ($\mu / \delta$)&\cellcolor{gray!25} \underline{97.6} / 3.4&\cellcolor{gray!25} \underline{97.9} / 3.2&\cellcolor{gray!25} \underline{98.7} / 3.2&\cellcolor{gray!25} \underline{92.7} / 21.8&\cellcolor{gray!25} \underline{97.5} / 3.8&\cellcolor{gray!25} \underline{92.1} / 8.0\\
& Adjacency & 97.2 & 98.4 & 96.4 & 98.8 & 98.6 & 95.8\\
& Incident  & 95.0 & 95.8 & 98.4 & 77.0 & 95.0 & 97.4\\
& Co-authorship    & 98.4 & 98.0 & 99.0 & 96.4 & 97.0 & 90.0 \\
& Friendship       & 98.4 & 98.0 & 99.4 & 92.8 & 98.0 & 89.4\\
& Social network   & 98.4 & 98.4 & 99.2 & 93.2 & 97.6 & 89.6\\
& Expert           & 98.2 & 99.0 & 99.6 & 97.8 & 98.8 & 90.6\\
\bottomrule
\end{tabular}
\caption{Comparison between our method and the baselines for various graph encoder functions based on their accuracy on different graph tasks using GPT-3.5 Turbo. The most effective prompting heuristic is underlined. The overall result is represented by its average~($\mu$) and an absolute difference~($\delta$) between its best and worst graph encoder.}
\label{table:our_method_main_table_gpt35}
\end{table*}


\section{Experiments}

\subsection{Dataset}
We conduct experiments on the GraphQA benchmark~\cite{fatemi2023talk}, which contains a set of diverse fundamental graph problems.
In this paper, there are 500 graphs in the training set and 500 graphs in the testing set. The graph size ranges from 5 to 20 nodes.

To enhance the diversity of graph structures for the basic graph tasks, we follow GraphQA to utilize the NetworkX library \citep{hagberg2008exploring} to generate random graphs using several algorithms, including the Erd\H{o}s-R\'enyi~(ER)~\cite{ErdosRenyi1959} model, Barabási–Albert model~(BA)~\cite{Albert_2002}, scale-free networks~(SFN)~\cite{doi:10.1126/science.286.5439.509} and stochastic block model~(SBM)~\citep{HOLLAND1983109} as well as generators for path, star, and complete graphs~\cite{fatemi2023talk}.

\subsection{Metrics}
We adopt exact match scores for evaluation, which measure whether the predicted answer matches the correct answer exactly.
For each task, we extract the answer from the LLM's response and compare it with the ground truth following the GraphQA benchmark. Unlike other tasks, \textit{connected nodes} requires LLMs to return a list of nodes. In this task, we do not consider the order of generated nodes. An answer is considered correct as long as the set of returned nodes is correct.


\subsection{Baselines}
Experiments are performed on four LLMs: GPT-3.5 Turbo~\cite{OpenAI2023GPT35Turbo}, Llama3-70B Instruct~\cite{dubey2024llama3herdmodels}, Mixtral-8x7B Instruct~\cite{jiang2024mixtralexperts}, and Mixtral-8x22B Instruct~\cite{Mixtral2024}. 
Then, we test them under both zero-shot~\cite{wei2022finetuned} and in-context learning (few-shot)~\cite{brown2020language} settings.
We also test Chain of Thought (CoT)~\cite{wei2023chainofthought} under the few-shot setting, allowing the LLMs to learn reasoning steps from sequences of examples and apply them to new inputs. 
In addition to the above baselines, results for the graph-based methods and transformer models baselines are in Appendix~\ref{appendix:baselines}.


\begin{table*}[t]

\centering

\resizebox{0.8\textwidth}{!}{

\setlength{\tabcolsep}{3pt}

\begin{tabular}{c|ccccccc}
\toprule
\textbf{Method} & \textbf{Generator} & \textbf{Edge Exist.} & \textbf{Node Deg.} & \textbf{Node Count} & \textbf{Edge Count} & \textbf{Conn. Nodes} & \textbf{Cycle Check}  \\ \midrule

\multirow{7}{*}{\begin{sideways}\method{adjacency}\end{sideways}}
& \cellcolor{gray!25} Overall ($\mu / \delta$)&\cellcolor{gray!25} 99.5 / 2.8&\cellcolor{gray!25} 96.8 / 11.2&\cellcolor{gray!25} 99.4 / 3.6&\cellcolor{gray!25} 99.0 / 2.2&\cellcolor{gray!25} 99.7 / 1.4&\cellcolor{gray!25} 98.7 / 4.2\\ 
& ER & 97.2  & 98.4 & 96.4 & 98.8 & 98.6 &  95.8\\
& BA & 100.0 & 98.2 & 100.0 & 99.8 & 100.0 & 100.0 \\
& SBM & 99.0  & 98.0 & 99.8 & 99.4 & 99.8 &  97.8\\
& Star & 100.0 & 97.0  & 99.8 & 97.8 & 100.0 & 99.2 \\
& SFN & 100.0 & 88.0 & 100.0 & 99.6 & 100.0 & 100.0 \\
& Path & 100.0 & 99.0 & 100.0 & 97.6 & 99.8 & 98.6 \\
& Complete & 100.0 & 99.2 & 100.0 & 99.8 & 99.8 & 99.4 \\
\midrule

\multirow{7}{*}{\begin{sideways}\method{friendship}\end{sideways}}
& \cellcolor{gray!25} Overall ($\mu / \delta$)&\cellcolor{gray!25} 99.7 / 1.6&\cellcolor{gray!25} 97.6 / 10.4&\cellcolor{gray!25} 99.7 / 0.8&\cellcolor{gray!25} 96.2 / 5.4&\cellcolor{gray!25} 99.4 / 2.0&\cellcolor{gray!25} 98.0 / 10.6\\
& ER & 98.4 & 98.0 & 99.4 & 92.8 & 98.0 &  89.4\\
& BA & 99.8 & 99.4 & 100.0 & 94.0 &  99.2 & 100.0 \\
& SBM & 99.8 & 99.6 & 99.4 & 97.4 & 99.2 & 97.6\\
& Star & 100.0 & 99.2 & 100.0 & 98.0 & 100.0 & 99.8\\
& SFN & 100.0 & 89.2 & 100.0 & 95.4 & 99.6 & 100.0 \\
& Path & 100.0 & 98.2 & 99.8 & 97.6 & 100.0 & 99.8 \\
& Complete & 99.8 & 99.4 & 99.2 & 98.2 & 100.0 & 99.6 \\
\bottomrule
\end{tabular}
}
\caption{Comparison of the accuracy of various graph generators on different graph tasks using two graph encoding functions with GPT-3.5 Turbo. The overall result is represented by its average~($\mu$) and an absolute difference~($\delta$) between its best and worst graph generator.}
\label{table:evaluation_graph_structure_robustness}
\end{table*}

\begin{table}[t]

\centering


\setlength{\tabcolsep}{3pt}

\begin{tabular}{c|cccc}
\toprule
\multirow{2}{*}{\textbf{Generator}} & \multicolumn{2}{c}{\textbf{Adjacency}} & \multicolumn{2}{c}{\textbf{Friendship}} \\
\cmidrule(lr){2-3} \cmidrule(lr){4-5}
& \textbf{Conn.} & \textbf{Degree} & \textbf{Conn.} & \textbf{Degree} \\ \midrule

\cellcolor{gray!25} Overall & \cellcolor{gray!25}99.7 & \cellcolor{gray!25}95.6 & \cellcolor{gray!25}99.3 & \cellcolor{gray!25}97.3 \\ 
BA & 99.8 & 98.0 & 99.6 & 99.8 \\
SBM & 99.2 & 97.8 & 99.8 & 99.4 \\
Star & 100.0 & 98.6 & 99.2 & 99.2 \\
SFN & 99.8 & 86.2 & 98.0 & 86.4 \\
Path & 100.0 & 94.8 & 99.4 & 99.2 \\
Complete & 99.2 & 98.2 & 99.8 & 99.6 \\
\bottomrule
\end{tabular}
\caption{ The generalization performance of \papertitle across various graph structures. The table shows the accuracy of \textit{Connected Nodes} and \textit{Node Degree} tasks for the adjacency and friendship graph encoding functions using GPT-3.5 Turbo. }
\label{table:generalization_cross_different_structure}
\end{table}

\subsection{Evaluation with Different Graph Encodings}
\label{sec:different_graph_encoding}
In this experiment, we evaluate the performance of \papertitle on the basic graph tasks: \textit{connected nodes}, \textit{cycle check}, \textit{edge existence}, \textit{edge count}, \textit{node degree} and \textit{node count} with different encoding functions including adjacency, incident, co-authorship, friendship, social network and expert encodings. Table \ref{table:our_method_main_table_gpt35} presents the accuracies for our method and baselines, evaluated using the GPT-3.5 Turbo.

As can be seen, the baselines perform poorly on nearly all basic graph tasks, especially on \textit{node degree}, \textit{edge count}, and \textit{connected nodes} tasks. In contrast, using code to encode solutions with our prompting technique significantly outperforms the baselines in all tasks. The results show more than 40\% improvements for most encoding methods in tasks \textit{node degree}, \textit{edge count}, and \textit{connected nodes} compared to all baselines. This demonstrates the potential of using code-based approaches to handle basic graph tasks effectively with LLMs.


We can also see that \papertitle shows robustness across various graph encoding functions on most basic graph tasks. For \textit{edge existence}, \textit{node degree}, \textit{node count}, and \textit{connected nodes}, the absolute difference between the best and worst graph encoding methods using \papertitle is less than 4\%. This consistent performance across different encoding methods validates the effectiveness of \papertitle.

\subsection{Evaluation with Different Graph Structures}
\label{sec:diff_graph_structure}
In \cref{sec:different_graph_encoding}, we conduct experiments solely on the graphs generated by the ER generator.
In this experiment, we evaluate the robustness of \papertitle across various graph structures. Here, various graph generators are employed to create graphs, including ER, BA, SBM, Star, SFN, Path, and Complete graphs~\cite{fatemi2023talk}. We sample 500 graphs for each graph structure. This experiment selects two graph encoding methods: friendship and adjacency. For each task, the LLM is provided with an exemplar that has the same graph structure as the test problem.

Table \ref{table:evaluation_graph_structure_robustness} shows our method's accuracy on different graph structures for all basic graph tasks. We can see that our method \papertitle performs robust reasoning in all graph structures. For example, the average accuracy across multiple graph structures in all tasks exceeds 90\%. For individual tasks, the accuracy differences between the best and worst performance using our approach with different graph structures are less than 5.4\%, including tasks like \textit{edge existence}, \textit{node count}, \textit{edge count}, and \textit{connected nodes}. This suggests that 
\papertitle is robust for the arithmetic tasks requiring the LLM to understand graph structure. 


\subsection{Evaluation with Generalization Across Diverse Graph Structures}
In this experiment, we test the ability of \papertitle to generalize across graph structures different from that of the exemplars. The difference between \cref{sec:diff_graph_structure} and this experiment is that we use an ER graph as the exemplar, and the test example is sampled using other graph structures, including BA, SBM, Star, SFN, Path, and Complete graphs. This is a more reflective real-world application to test LLMs' ability to extrapolate and enhance their basic knowledge when faced with more complex graph-structured data. This experiment tests whether the LLMs can generate the correct code when provided with exemplars from a different graph structure. Also, we consider the graph encoding methods of friendship and adjacency in the experiment.

As shown in Table \ref{table:generalization_cross_different_structure}, the average accuracy for the tasks remained high across various graph structures despite the shift from ER to other more complex graph types for testing. For both graph encoding functions, the average accuracies are higher than 95\%, demonstrating robustness for the two tasks' performance. When examining graph structures, our method can generalize effectively from an ER graph structure to a more complex one and maintain high accuracy across different tasks. This experiment demonstrates that our method \papertitle owns robustness and generalization across different graph structures for most basic graph tasks.


\begin{table*}[t]
\centering
 \small
\setlength{\tabcolsep}{3pt}
\begin{tabular}{c|c|cccccc}
\toprule
\textbf{Method} & \textbf{Encoding} & \textbf{Edge Exist.} & \textbf{Node Deg.} & \textbf{Node Count} & \textbf{Edge Count} & \textbf{Conn. Nodes} & \textbf{Cycle Check} \\ 

\midrule
\multirow{3}{*}{  \begin{tabular}{@{}c@{}}
    \method{\footnotesize Llama3-70B} \\
    \method{\footnotesize (zero-shot)}
  \end{tabular}}
  
&  \cellcolor{gray!25} Overall ($\mu / \delta$) & \cellcolor{gray!25} 92.3 / 4.6 & \cellcolor{gray!25} 79.4 / 4.8 & \cellcolor{gray!25} \underline{84.2} / 31.6 & \cellcolor{gray!25} 33.9 / 5.4 & \cellcolor{gray!25} 71.5 / 8.6 & \cellcolor{gray!25} \underline{94.0} / 1.6 \\
& Adjacency & 90.0 & 77.0 & 100.0 & 31.2 & 75.8 & 94.8 \\
& Friendship & 94.6 & 81.8 & 68.4 & 36.6 & 67.2 & 93.2 \\

\midrule
\multirow{3}{*}{  \begin{tabular}{@{}c@{}}
    \method{\footnotesize Llama3-70B} \\
    \method{\footnotesize (our)}
  \end{tabular}}
  
&  \cellcolor{gray!25} Overall ($\mu / \delta$) & \cellcolor{gray!25} \underline{98.4} / 0.8 & \cellcolor{gray!25} \underline{89.6} / 5.2 & \cellcolor{gray!25} 81.8 / 14.0 & \cellcolor{gray!25} \underline{98.0} / 1.2 & \cellcolor{gray!25} \underline{90.8} / 11.2 & \cellcolor{gray!25} 93.0 / 2.4 \\
& Adjacency & 98.0 & 87.0 & 74.8 & 97.4 & 96.4 & 91.8 \\
& Friendship & 98.8 & 92.2 & 88.8 & 98.6 & 85.2 & 94.2 \\
\midrule
\multirow{3}{*}{  \begin{tabular}{@{}c@{}}
    \method{\footnotesize Mixtral-8x22B} \\
    \method{\footnotesize (zero-shot)}
  \end{tabular}}
  
& \cellcolor{gray!25} Overall ($\mu / \delta$) & \cellcolor{gray!25} 93.2 / 2.0 & \cellcolor{gray!25} 70.8 / 2.8 & \cellcolor{gray!25} \underline{97.3} / 0.2 & \cellcolor{gray!25} 50.9 / 1.8 & \cellcolor{gray!25} 52.4 / 12.8 & \cellcolor{gray!25} \underline{80.2} / 7.2 \\
& Adjacency & 94.2 & 72.2 & 97.2 & 50.0 & 58.8 & 83.8 \\
& Friendship & 92.2 & 69.4 & 97.4 & 51.8 & 46.0 & 76.6 \\

\midrule
\multirow{3}{*}{  \begin{tabular}{@{}c@{}}
    \method{\footnotesize Mixtral-8x22B} \\
    \method{\footnotesize (our)}
  \end{tabular}}
  
& \cellcolor{gray!25} Overall ($\mu / \delta$) & \cellcolor{gray!25} \underline{94.7} / 6.2 & \cellcolor{gray!25} \underline{80.9} / 2.2 & \cellcolor{gray!25} 97.0 / 2.4 & \cellcolor{gray!25} \underline{79.9} / 3.0 & \cellcolor{gray!25} \underline{79.5} / 3.8 & \cellcolor{gray!25} 76.2 / 15.2 \\
& Adjacency & 91.6 & 79.8 & 95.8 & 81.4 & 77.6 & 83.8 \\
& Friendship & 97.8 & 82.0 & 98.2 & 78.4 & 81.4 & 68.6 \\
\bottomrule
\end{tabular}
\caption{Comparison between our method and zero-shot baseline for the two graph encoder functions based on their accuracy in different graph tasks using Llama3-70B Instruct and Mixtral-8x22B Instruct.
The most effective prompting heuristic is underlined. 
The overall result is represented by its average~($\mu$) and an absolute difference~($\delta$) between its best and worst graph encoder.}
\label{table:generalization_cross_different_models}
\end{table*}

\begin{table*}[t]
\centering
\small
\setlength{\tabcolsep}{3pt}
\begin{tabular}{c|c|cccccc}
\toprule
\textbf{Method} & \textbf{Encoding} & \textbf{Edge Exist.} & \textbf{Node Deg.} & \textbf{Node Count} & \textbf{Edge Count} & \textbf{Conn. Nodes} & \textbf{Cycle Check} \\ 

\midrule
\multirow{3}{*}{  \begin{tabular}{@{}c@{}}
    \method{\footnotesize Mixtral-8x7B} \\
    \method{\footnotesize (1-shot)}
  \end{tabular}}
& \cellcolor{gray!25} Overall ($\mu / \delta$) & \cellcolor{gray!25} \underline{33.6} / 26.4 & \cellcolor{gray!25} 26.6 / 11.6 & \cellcolor{gray!25} \underline{59.2} / 1.6 & \cellcolor{gray!25} 61.4 / 18.0 & \cellcolor{gray!25} 38.6 / 8.4 & \cellcolor{gray!25} 25.8 / 10.8 \\

& Adjacency & 46.8 & 20.8 & 58.4 & 70.4 & 34.4 & 20.4 \\
& Friendship & 20.4 & 32.4 & 60.0 & 52.4 & 42.8 & 31.2 \\

\midrule
\multirow{3}{*}{  \begin{tabular}{@{}c@{}}
    \method{\footnotesize Mixtral-8x7B} \\
    \method{\footnotesize (2-shot)}
  \end{tabular}}

& \cellcolor{gray!25} Overall ($\mu / \delta$)&\cellcolor{gray!25} 10.6 / 8.4&\cellcolor{gray!25} 47.4 / 6.8&\cellcolor{gray!25} 50.6 / 13.2&\cellcolor{gray!25} 68.8 / 0.8&\cellcolor{gray!25} 53.0 / 34.8&\cellcolor{gray!25} 32.2 / 17.2\\

& Adjacency & 14.8 & 44.0 & 57.2 & 68.4 & 35.6 & 23.6 \\
& Friendship & 6.4 & 50.8 & 44.0 & 69.2 & 70.4 & 40.8 \\

\midrule
\multirow{3}{*}{  \begin{tabular}{@{}c@{}}
    \method{\footnotesize Mixtral-8x7B} \\
    \method{\footnotesize (3-shot)}
  \end{tabular}}

& \cellcolor{gray!25} Overall ($\mu / \delta$) & \cellcolor{gray!25} 8.2 / 2.8 & \cellcolor{gray!25} \underline{59.4} / 14.0 & \cellcolor{gray!25} 46.6 / 9.2 & \cellcolor{gray!25} \underline{76.4} / 0.0 & \cellcolor{gray!25} \underline{56.6} / 28.4 & \cellcolor{gray!25} \underline{53.4} / 25.2 \\

& Adjacency & 9.6 & 66.4 & 51.2 & 76.4 & 42.4 & 40.8 \\
& Friendship & 6.8 & 52.4 & 42.0 & 76.4 & 70.8 & 66.0 \\
\bottomrule
\end{tabular}

\caption{Comparison of various graph encoder functions based on their accuracy on different graph tasks using Mixtral-8x7B Instruct. The table shows performance for one-shot, two-shot, and three-shot setups, highlighting the results for its average~($\mu$) and an absolute difference~($\delta$) between its best and worst graph encoder. The most effective prompting heuristic is underlined. }
\label{table:scalability_mixtral-8x7B}
\end{table*}

\begin{figure}[t]
    \centering
    \includegraphics[width=\columnwidth]{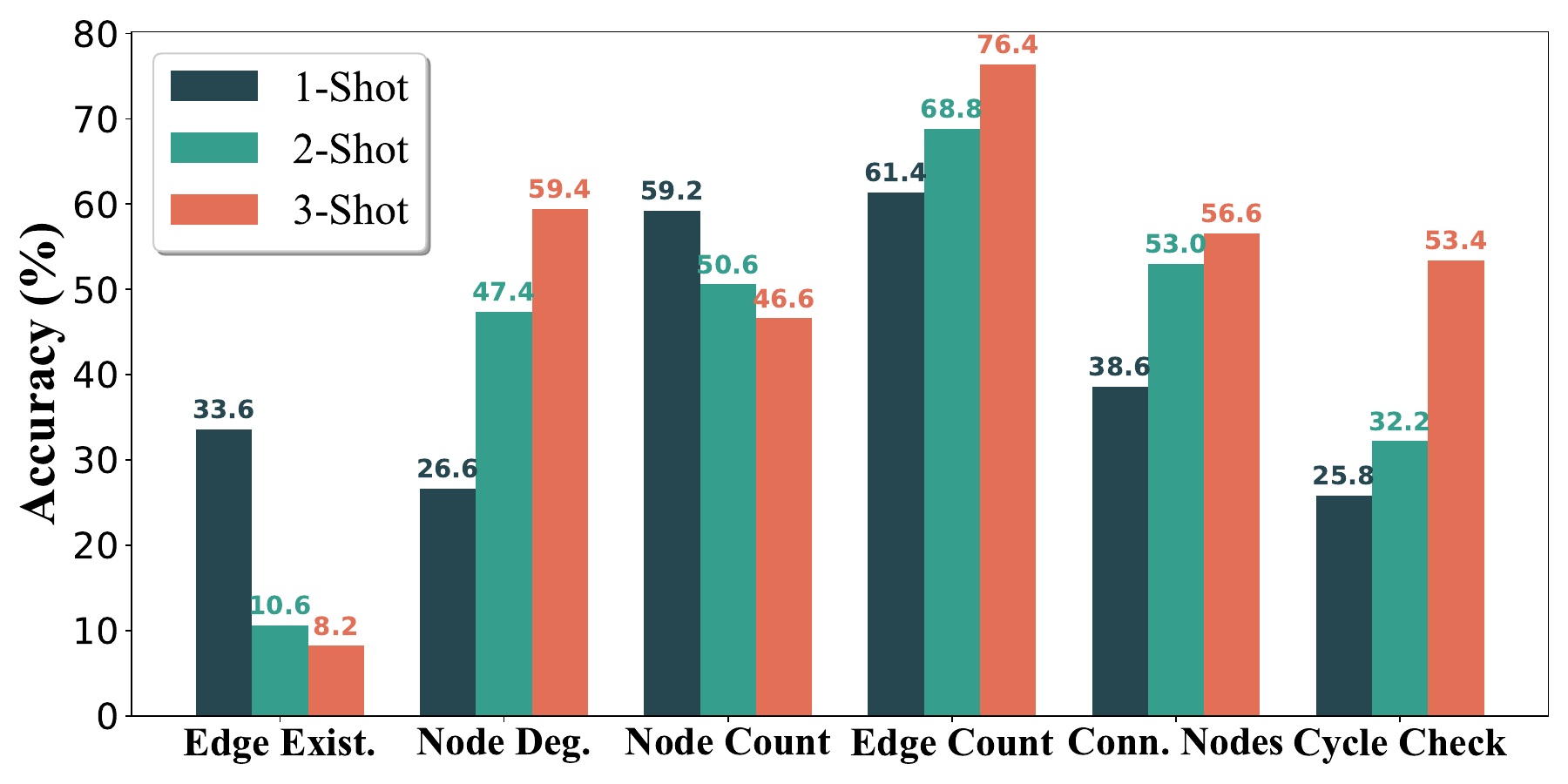}
    \caption{Impact of providing one, two, or three examples on the overall performance of the Mixtral-8x7B Instruct model across various graph tasks.}
    \label{fig:histogram_x87B_few_shots}
\end{figure}

\subsection{Evaluation with Different LLMs}
In this experiment, we investigated whether 
\papertitle can be applied to other LLMs. We selected Llama3-70B Instruct and Mixtral-8x22B Instruct and followed the same experiment design in the first experiment (\cref{sec:different_graph_encoding}). We evaluate models on various graph tasks with the ER generator and two graph encoding functions. We aimed to assess the robustness and effectiveness of \papertitle across different LLMs.


As a result in \cref{table:generalization_cross_different_models}, using \papertitle for Llama3-70B Instruct and Mixtral-8x22B Instruct outperforms zero-shot approaches on most tasks, including the \textit{edge existence}, \textit{node degree}, \textit{edge count} and \textit{connected nodes} tasks. The improvements in average accuracy score range from 6.1\% to 64.1\% for Llama3-70B Instruct and from 1.5\% to 29.0\% for Mixtral-8x22B Instruct. The experimental results show that our framework is robust and effective across LLMs with varying coding abilities.

\subsection{Evaluation with Smaller LLM and Weaker Coding Capability}

In this experiment, we evaluate our method on a smaller LLM: Mixtral-8x7B Instruct, whose coding ability is much lower than GPT-3.5 Turbo. For example, Mixtral-8x7B Instruct only achieves a 39.6\% pass rate on HumanEval~\cite{chen2021evaluating}, while GPT-3.5 Turbo achieves 70.7\%. Also, we aimed to investigate whether increasing the number of examples in \papertitle could boost the performance of smaller LLMs with more limited coding capability. We follow the same few-shot set-up of previous experiments and also increase the exemplar numbers to two and three.
 
As illustrated in Figure~\ref{fig:histogram_x87B_few_shots}, we can first see that the performance of Mixtral-8x7B Instruct is much lower than GPT-3.5 Turbo, Llama3-70B Instruct, and Mixtral-8x22B Instruct. Then, the Mixtral-8x7B Instruct model exhibited varied improvements across some tasks when provided with more examples. Specifically, the average accuracy for \textit{node degree}, \textit{edge count}, \textit{cycle check}, and \textit{connected nodes} tasks increased by 20.8\%, 7.4\%, 14.4\%, and 6.4\%, respectively, when we increase the exemplar number from one-shot to two-shot. This upward trend persisted with the three-shot setup, indicating a consistent enhancement in performance across these graph tasks. However, for tasks such as \textit{edge existence} and \textit{node count}, the average accuracy decreased by 23.0\% and 8.6\%, respectively, when moving from the one-shot to the two-shot setup. This suggests an increased sensitivity to exemplars, as pointed out by~\cite{chen2023program}. 

In short, a smaller LLM with limited coding ability may struggle to learn effectively from the exemplar, and we can improve its performance on most tasks by adding more exemplars.

\section{Conclusion}
In this work, we present the first study on enhancing the graph reasoning capabilities of LLMs using code for basic graph tasks. Previous works usually convert graphs to natural language. Here, we demonstrate that our proposed method can use programming languages to significantly improve performance across various basic graph tasks. The experiments consistently show robustness across different graph encoding functions and various graph structures, highlighting the effectiveness of our approach. We also illustrate that this improvement in graph reasoning can be extended to other LLMs. Then, we also test our method with Mixtral-8x7B Instruct to show the performance of our method with less powerful coding capabilities. These novel results provide valuable insights into the graph reasoning abilities of LLMs using code, which can boost performance on basic graph tasks by 1.3\% to 58.6\% on different tasks. 

In short, our work provides a new direction for using LLMs to tackle basic graph tasks, inspiring further exploration of using the code ability of LLMs to tackle other graph problems~\cite{hu2023privacy, bai2023knowledgegraphreasoningentities, hu2402fedcqa}. For future works, with the recent progress of LLMs~\cite{du2024llms}, we can test more LLMs, like Gemma 2~\cite{gemmateam2024gemma2improvingopen} and Claude 3.5~\cite{claude3.5}. Meanwhile, we can augment our methods with more decoding methods~\cite{choi2023kcts} to generate code with higher quality. Also, we can add various kinds of knowledge to help the understanding of graph questions, including commonsense knowledge~\cite{sap2019atomic, deng2023gold, do2024constraintchecker, wu2023commonsense, shen2022dense}, discourse knowledge~\cite{wang2023cola, wang2022subeventwriter, fang2024getting, cui2024exploring, cui2024odyssey}, abstraction knowledge~\cite{wang2024abspyramid, wang2024absinstruct} and others~\cite{wang2022legal, zheng2024knowcomp, hu2023independent, bai2024advancingabductivereasoningknowledge}. We also can use multimodal information to help LLMs better understand the tasks~\cite{shen2024vcd, zong2023tilfa}.

\section*{Acknowledgements}
The authors of this paper were supported by the NSFC Fund (U20B2053) from the NSFC of China, the RIF (R6020-19 and R6021-20) and the GRF (16211520 and 16205322) from RGC of Hong Kong. We also thank the support from the Tencent AI Lab Rhino-Bird Focused Research Program and the UGC Research Matching Grants (RMGS20EG01-D, RMGS20CR11, RMGS20CR12, RMGS20EG19, RMGS20EG21, RMGS23CR05, RMGS23EG08).



\bigskip

\newpage
\bibliography{aaai25}

\newpage

\newpage

\appendix
\section{Task Descriptions for Basic Graph Tasks}
\label{appendix:task_description}

We provide a complete task description for the edge existence task as an example in Table~\ref{table:task_description_complete_prompt}. The task descriptions for various tasks within the prompt are provided as follows:

\begin{itemize}
\item \textbf{Connected Nodes}: List all the nodes connected to a specific node in a graph.
\item \textbf{Cycle Check}: Determine whether a graph contains any cycles.
\item \textbf{Edge Existence}: Determine whether there is a direct connection(edge) between two specified nodes in a graph.
\item \textbf{Edge Count}: Determine the total number of edges in a graph.
\item \textbf{Node Degree}: Calculate the number of edges connected to a specific node in a graph.
\item \textbf{Node Count}: Calculate the number of nodes in a graph.
\end{itemize}

We explore these tasks because they are the basis for building more complex graph tasks. For instance, finding the minimum distance between two nodes (\textit{shortest path}) relies on exploring all possible paths (\textit{connected nodes}). Determining clusters of densely \textit{connected nodes} (community detection) relies on the presence of cycles since communities in a network form cycles or closely-knit loops.

\begin{table}[!t]
\centering
\small
\caption{Complete task description of CodeGraph for the \textit{Edge existence} task.} 
\begin{tabular}{p{0.95\linewidth}}
\toprule
\textbf{Complete Task Description of \papertitle}
\\
\midrule
You are tasked with assisting a user who is seeking help with graph networks and Python programming. The user is looking for guidance from an AI that is knowledgeable in graph networks, proficient in Python, and capable of providing bug-free solutions to graph network problems. You will first be given an example to learn about the way to answer. Then you should adhere strictly to the provided program in the example to answer the graph network question.\\
For this task, please determine whether there is an an edge between node i and node j in the undirected graph G.\\
Write a piece of Python code to return the answer in a variable \texttt{`ans'}. Please enclose the code with \texttt{\# CODE START} and \texttt{\# CODE END}. Assume \texttt{`edges'} is an empty list (\texttt{edges = []}) if not provided.\\
\bottomrule
\label{table:task_description_complete_prompt}
\end{tabular}
\end{table}

\section{Graph Encoding Functions}
\label{appendix:graph_encoding_function}

We follow the work \citet{fatemi2023talk} to construct graph encoding functions for representing textual descriptions of graphs.

\begin{itemize}
\item \textbf{Co-authorship}: Common English first names are used for node encoding, and edges indicate co-authorship connections.
\end{itemize}
\noindent\begin{center}
\noindent\fbox{%
\parbox{\dimexpr\columnwidth-0.5cm\relax}{
\textbf{Co-authorship:}\\ 
G describes a co-authorship graph among James, \\
Robert, John, Michael, and David.\\
In this co-authorship graph:\\
James and Robert wrote a paper together.\\
James and John wrote a paper together.\\
Robert and John wrote a paper together.\\
John and Michael wrote a paper together.\\
John and David wrote a paper together.
}
}
\end{center}

\begin{itemize}
\item \textbf{Incident}: Nodes are encoded as integers, and edges are represented by their incident connections.
\end{itemize}
\noindent\begin{center}
\fbox{%
\parbox{\dimexpr\columnwidth-0.5cm\relax}{
\textbf{Incident:}\\ 
G describes a graph among 0, 1, 2, 3, and 4.\\
In this graph:\\
Node 0 is connected to nodes 1, 2.\\
Node 1 is connected to nodes 0, 2.\\
Node 2 is connected to nodes 0, 1, 3, 4\\
Node 3 is connected to nodes 2\\
Node 4 is connected to node 2.
}
}
\end{center}

\begin{itemize}
\item \textbf{Social Network}: Nodes are encoded with familiar English first names, and edges illustrate social network ties.
\end{itemize}
\noindent\begin{center}
\fbox{%
\parbox{\dimexpr\columnwidth-0.5cm\relax}{
\textbf{Social Network:}\\ 
G describes a social network graph among James, Robert, John, Michael, and David.\\
We have the following edges in G:\\
James and Robert are connected.\\
James and John are connected.\\
Robert and John are connected.\\
John and Michael are connected.\\
John and David are connected.
}
}
\end{center}

\begin{itemize}
\item \textbf{Expert}: Nodes are encoded as alphabet letters, and edges are represented by arrows.
\end{itemize}
\noindent\begin{center}
\fbox{%
\parbox{\dimexpr\columnwidth-0.5cm\relax}{
\textbf{Expert:}\\ 
You are a graph analyst and you have been given a graph G among A, B, C, D, and E.
G has the following undirected edges:\\
A -$>$ B\\
A -$>$ C\\
B -$>$ C\\
C -$>$ D\\
C -$>$ E
}
}
\end{center}

\section{Concrete Prompt for Graph Tasks}
\label{appendix:graph_questions}
We present the specific prompts for various graph tasks as follows:

\begin{itemize}
\item \textbf{Connected Nodes}: 

\texttt{Q: List all the nodes connected to 
`David' in alphabetical order.}

\item \textbf{Cycle Check}: 

\texttt{Q: Is there a cycle in this graph?}

\item \textbf{Edge Existence}: 

\texttt{Q: Is node 0 connected to node 1?}

\item \textbf{Edge Count}: 

\texttt{Q: How many edges are in this graph?}

\item \textbf{Node Degree}: 

\texttt{Q: What is the degree of node 2?}

\item \textbf{Node Count}:  

\texttt{Q: How many nodes are in this graph?}
\end{itemize}

\section{Sample Code}
\label{appendix:sample_programs}

We provided our sample code for each task as follows:

\begin{itemize}
    \item Node Count
\end{itemize}
\begin{tcolorbox}[
    enhanced,
    boxrule=0.5pt,
    colback=white!10, 
    colframe=blue!75!black, 
    width=\columnwidth, 
    sharp corners, 
    boxsep=5pt, 
    left=2pt, right=2pt, top=2pt, bottom=2pt, 
    listing only
]
\begin{lstlisting}
# CODE START
nodes = "<node_list>"
def count_nodes(nodes):
    return len(nodes)
ans = count_nodes(nodes)
# CODE END
\end{lstlisting}
\end{tcolorbox}

\begin{itemize}
    \item Edge Count
\end{itemize}
\begin{tcolorbox}[
    enhanced,
    boxrule=0.5pt,
    colback=white!10, 
    colframe=blue!75!black, 
    width=\columnwidth, 
    sharp corners, 
    boxsep=5pt, 
    left=2pt, right=2pt, top=2pt, bottom=2pt, 
    listing only
]
\begin{lstlisting}
# CODE START
from typing import List, Tuple
def count_edges(edges: List[Tuple[str, str]]) -> int:
    unique_edges = set()
    for u, v in edges:
        edge = tuple(sorted((u, v)))
        unique_edges.add(edge)
    return len(unique_edges)
edges = "<edge_list>"
ans = count_edges(edges)
# CODE END
\end{lstlisting}
\end{tcolorbox}

\begin{itemize}
    \item Edge Existence
\end{itemize}
\begin{tcolorbox}[
    enhanced,
    boxrule=0.5pt,
    colback=white!10, 
    colframe=blue!75!black, 
    width=\columnwidth, 
    sharp corners, 
    boxsep=5pt, 
    left=2pt, right=2pt, top=2pt, bottom=2pt, 
    listing only
]
\begin{lstlisting}
# CODE START
source = "<source_node>"
target = "<target_node>"
edges =  "<edge_list>"
def edge_existence(edges, source, target):
    edges_set = set()
    for u, v in edges:
        edges_set.add((u,v))
        edges_set.add((v,u))
    if (source, target) in edges_set: return True
    else: return False
ans = edge_existence(edges,source,target)
# CODE END
\end{lstlisting}
\end{tcolorbox}

\begin{figure}[!ht]
\begin{itemize}
    \item Node Degree
\end{itemize}

\begin{tcolorbox}[
    enhanced,
    boxrule=0.5pt,
    colback=white!10, 
    colframe=blue!75!black, 
    width=\columnwidth, 
    sharp corners, 
    boxsep=5pt, 
    left=2pt, right=2pt, top=2pt, bottom=2pt, 
    listing only
]
\begin{lstlisting}
# CODE START
from collections import defaultdict
from typing import Set, Dict, List, Tuple
def get_adjacency_list(edges: List[Tuple[str, str]]) -> Dict[str, Set[str]]:
    adjacency = defaultdict(set)
    for each_edge in edges:
        u, v = each_edge
        adjacency[u].add(v)
        adjacency[v].add(u)
    return adjacency
def get_node_degree(target_node: str, adjacency_list: Dict[str, Set[str]]) -> int:
    return len(adjacency_list[target_node])
edges = "<edge_list>"
adjacency_list = get_adjacency_list(edges)
target_node = '<target_node>'
ans = get_node_degree(target_node, adjacency_list)
# CODE END
\end{lstlisting}
\end{tcolorbox}
\end{figure}

\begin{figure}[!ht]
\begin{itemize}
    \item Cycle Check
\end{itemize}
\begin{tcolorbox}[
    enhanced,
    boxrule=0.5pt,
    colback=white!10, 
    colframe=blue!75!black, 
    width=\columnwidth, 
    sharp corners, 
    boxsep=5pt, 
    left=2pt, right=2pt, top=2pt, bottom=2pt, 
    listing only
]
\begin{lstlisting}
# CODE START
def has_cycle(graph):
    visited = set()
    for node in graph:
        if node not in visited:
            if dfs(graph, node, visited, None):
                return True
    return False
def dfs(graph, node, visited, parent):
    visited.add(node)
    for neighbor in graph[node]:
        if neighbor not in visited:
            if dfs(graph, neighbor, visited, node):
                return True
        elif neighbor != parent:
            return True
    return False
nodes = "<node_list>"
edges = "<edge_list>"
graph = {node: [] for node in nodes}
for edge in edges:
     graph[edge[0]].append(edge[1])
if has_cycle(graph):
    ans = "Has cycle."
else:
    ans = "No cycle."
# CODE END
\end{lstlisting}
\end{tcolorbox}
\end{figure}

\begin{figure}[!ht]
\begin{itemize}
    \item Connected Nodes
\end{itemize}

\begin{tcolorbox}[
    enhanced,
    boxrule=0.5pt,
    colback=white!10, 
    colframe=blue!75!black, 
    width=\columnwidth, 
    sharp corners, 
    boxsep=5pt, 
    left=2pt, right=2pt, top=2pt, bottom=2pt, 
    listing only
]
\begin{lstlisting}
# CODE START
from collections import defaultdict
from typing import Set, Dict, List, Tuple
def get_adjacency_list(edges: List[Tuple[str, str]]) -> Dict[str, Set[str]]:
    adjacency = defaultdict(set)
    for each_edge in edges:
        u, v = each_edge
        adjacency[u].add(v)
        adjacency[v].add(u)
    return adjacency
def get_connected_nodes(target_node: str, adjacency_list: Dict[str, Set[str]]) -> str:
    if target_node in adjacency_list and adjacency_list[target_node]:
        connected_nodes = sorted(adjacency_list[target_node], key=lambda x: (x.isdigit(), int(x) if x.isdigit() else x))
        return ', '.join(connected_nodes)
    else:
        return "No nodes"
edges = "<edge_list>"
adjacency_list = get_adjacency_list(edges)
target_node = '<target_node>'
ans = get_connected_nodes(target_node, adjacency_list)
# CODE END
\end{lstlisting}
\end{tcolorbox}
\end{figure}

\newpage
\section{Supplementary Baselines}
\label{appendix:baselines}

To thoroughly assess the performance of \papertitle in graph reasoning tasks, we benchmark it against three categories of baselines:  graph-based approaches, transformer models, and prompting-based approaches.

\begin{enumerate}

    \item \textbf{Graph-based methods.} Graph-based methods are trained in a task-specific manner and directly use graph-structured data as input. These methods utilize the graph's topology (the connections between nodes) to learn representations, which are then used to make predictions. We copy the results from the original papers, including Message Passing Neural Network (MPNN)~\cite{gilmer2017neuralmessagepassingquantum}, Graph Convolutional Networks (GCN)~\cite{kipf2017semisupervisedclassificationgraphconvolutional}, Graph Isomorphism Network (GIN)~\cite{xu2019powerfulgraphneuralnetworks} and GraphToken~\cite{perozzi2024letgraphtalkingencoding}. GraphToken provides tokens generated from Graph Neural Networks (GNNs) and integrates them with textual tokens as soft-tokens inputs to an LLM.

    \item \textbf{Transformer models.} The task-specific vanilla transformer models~\cite{vaswani2023attentionneed} were directly trained or fine-tuned on different scales of the GraphQA training set. These models include the 60M transformer-1K, 60M transformer-100K, 11B transformer (FT)-1K, and PaLM 2~\cite{anil2023palm2technicalreport}
    transformers of sizes XXS and XS. The suffix in the method names in the table below indicates the number of training examples used from the GraphQA training set, with "FT" signifying fine-tuning. We report the results of these models as presented in the original paper.
        
    \item \textbf{Prompting-based methods \textbf{(Ours)}.} These methods provide an LLM with a task description, as well as textual descriptions of the graph and the question within the prompt. We present the following approaches and their corresponding results in this paper.
    \begin{itemize}
        \item \method{\normalsize Zero-shot:} In this approach, the input to the model includes textual descriptions of a graph and a question for the task without providing extra examples or demonstrations.
        \item \method{\normalsize Few-Shot:} In this approach, few-shot learning~\cite{Brown2020LanguageMA} provides a few additional question-and-answer pairs in the prompt, enabling the LLM to learn from them and handle a test example during inference.
        \item \method{\normalsize Chain of Thought (CoT):} Chain of Thought~\cite{wei2023chainofthought} provides intermediate steps to solve the task, enabling an LLM to learn and apply these reasoning processes to new inputs.
        \item \method{\normalsize \textbf{CodeGraph}}: Our method guides the LLM to generate a Python program that solves the task. The LLM translates the textual descriptions of the graphs into code, storing the nodes and/or edges as input parameters for a Python function provided by exemplars, depending on the task. The program is then executed by an external interpreter to produce the final answer.
    \end{itemize}

\end{enumerate}

In Table~\ref{table:supplementary_baselines}, the results for MPNN, GCN, and GIN, as well as the results for transformer models, are taken from~\cite{sanford2024understandingtransformerreasoningcapabilities}. The results for GraphToken are taken from~\cite{perozzi2024letgraphtalkingencoding}. For the prompting-based methods, the results are taken from the overall average accuracies for various tasks in Table~\ref{table:our_method_main_table_gpt35}.

\begin{table*}[t]
\resizebox{\textwidth}{!}{%
\setlength{\tabcolsep}{3pt}
\begin{tabular}{llcccccc} 
\toprule
 & \textbf{Method} & \textbf{Node count} & \textbf{Edge count} & \textbf{Edge existence} & \textbf{Node degree} &  \textbf{Connected nodes} & \textbf{Cycle check} \\ \midrule

\multirow{4}{*}{\begin{sideways}\scriptsize Graph-based\end{sideways}}  & GCN & 6.4 & 1.2 & 47.0 & 9.8 & N/A & 83.2 \\
& MPNN & 19.4 & 16.2 & 69.2 & \textbf{99.4} & N/A & \textbf{99.0} \\
& GIN & 71.2 & 4.4 & 71.2 & 36.2 & N/A & \underline{98.8} \\
& GraphToken & \underline{99.6} & 42.6 & 73.8 & 96.2 & 26.4 & 95.6 \\
\midrule

\multirow{5}{*}{\begin{sideways}\scriptsize Transformers\end{sideways}} &
60M transformer-1K
& \textbf{100.0} & \textbf{100.0} &	67.6 &	31.5 &	N/A &	97.1 \\
& 
60M transformer-100K
& \textbf{100.0} &	\textbf{100.0} & 96.1 &	91.7 &	N/A &	98.0 \\

& XXS transformer (FT)-1K & \textbf{100.0} & 70.6 & 73.0 & 31.0 & N/A & 98.0 \\
& XS transformer (FT)-1K & \textbf{100.0} & 73.2 & \underline{98.6} & 50.6 & N/A & 96.8 \\
&
11B transformer (FT)-1K
& \textbf{100.0} & 45.0 & \textbf{100.0} & 68.8 & N/A & 98.0 \\
\midrule

\multirow{4}{*}{\begin{sideways}\scriptsize Ours \end{sideways}} 

& \acr{zero-shot} & 97.4 & 34.1 & 72.9 & 49.5 & 40.6 & 85.3 \\

& \acr{One-shot} & 85.6 & 28.3 & 71.8 & 44.6 & 36.5 & 86.9 \\

& \acr{cot(one-shot)} & 81.2 & 26.1 & 72.5 & 42.8 & 37.2 & 82.6\\

& \acr{\textbf{codegraph(one-shot)}} & 98.7 & \underline{92.7} & 97.6 & \underline{97.9} & \textbf{97.5} & 92.1 \\

\bottomrule
\end{tabular}
}
\caption{
Comparison of various approaches on the basic graph tasks of GraphQA. The transformers are either directly trained or fine-tuned on different sizes of the GraphQA training set, using 1K or 100K examples. `N/A' indicates that the results for the \textit{connected nodes} task are not available in the paper~\cite{sanford2024understandingtransformerreasoningcapabilities}. 
The highest accuracy for each task is highlighted in bold, while the second highest is underlined.
}\label{table:supplementary_baselines}

\end{table*}



\section{Implementation Details}
\label{appendix:implementation}
In our work, we utilized the Azure OpenAI API\footnote{Azure OpenAI API: \url{https://learn.microsoft.com/en-us/azure/ai-services/openai/}} and the Deepinfra API\footnote{Deepinfra API: \url{https://deepinfra.com}} for our experiments. For the LLMs, we tested \textit{GPT-3.5-Turbo (0301)},  \textit{Llama3-70B Instruct}~\cite{dubey2024llama3herdmodels}, \textit{Mixtral-8x22B Instruct}~\cite{mixtral8x22b_instruct_hf}, and \textit{Mixtral-8x7B Instruct}~\cite{jiang2023mistral}. The decoding temperatures for GPT-3.5 and Llama3-70B Instruct were set to 0.7, and for Mixtral models, it was set to 1.0. We set other parameters, such as probability mass (Top\_p), frequency penalty, and presence penalty, to the default values.

For each of the ER, BA, SFN, SBM, Path, Star, and Complete graphs, we sample 500 examples for both the training and testing sets. The training graphs were utilized to construct examples for exemplars in few-shot learning, while the testing graphs were used to pose questions that we expected  LLMs to answer.

In detail, the graph generator accepts specific parameters for each algorithm to produce diverse graphs. The ER graphs' probability of creating an edge is randomly sampled between 0 and 1. For the BA graphs, the number of edges attached from a new node is randomly chosen between 1 and the number of nodes minus 1. For the SBM graphs, the number of communities varies from 2 to 10.

\section{Examples of \papertitle Prompts for Various Graph Tasks }
We provide these \papertitle prompts for the basic graph tasks in GraphQA to improve reproducibility. Please note that the following prompts are adapted using the adjacency graph encoding function, which can be changed to other encodings, such as co-authorship or incident graph encoding functions. The prompts are listed as follows: Tables 
\ref{table:codegraph_node_count_prompt},
\ref{table:codegraph_edge_count_prompt},
\ref{table:codegraph_edge_existence_prompt},
\ref{table:codegraph_node_degree_prompt},
\ref{table:codegraph_connected_nodes_prompt},
and
\ref{table:codegraph_cycle_check_prompt}.

\begin{table*}[!t]
\centering
\small
\caption{An example of \papertitle prompting with the \textit{Adjacency} graph encoding function on the \textit{Node count} task.} 
\begin{tabular}{p{0.95\linewidth}}
\toprule
\textbf{CodeGraph Prompts for the \textit{Node count} Task (1-Shot)}
\\
\midrule
You are tasked with assisting a user who is seeking help with graph networks and Python programming. The user is looking for guidance from an AI that is knowledgeable in graph networks, proficient in Python, and capable of providing bug-free solutions to graph network problems. You will first be given an example to learn about the way to answer. Then you should adhere strictly to the provided program in the example to answer the graph network question.\\
For this task, please count the number of nodes in the undirected graph G.\\
Write a piece of Python code to return the answer in a variable \texttt{`ans'}. Please enclose the code with \texttt{\# CODE START} and \texttt{\# CODE END}. Assume \texttt{nodes} is an empty list (\texttt{nodes = []}) if not provided.\\
\textbf{Q}: How many nodes are in this graph? \\
In an undirected graph, (i,j) means that node i and node j are connected with an undirected edge. G describes a graph among nodes 0, 1, 2, 3, 4, and 5. \\
The edges in G are: (0, 2) (1, 2) (1, 4) (1, 5).\\
\textbf{A}: \begin{verbatim}
# CODE START
nodes = ['0','1','2','3','4','5']
def count_nodes(nodes):
    return len(nodes)
ans = count_nodes(nodes)
# CODE END
\end{verbatim}
\\
Write a piece of Python code to return the answer in a variable \texttt{`ans'}. Please enclose the code with \texttt{\# CODE START} and \texttt{\# CODE END}. Assume \texttt{nodes} is an empty list (\texttt{nodes = []}) if not provided.\\
\textbf{Q}: How many nodes are in this graph?  \\
In an undirected graph, (i,j) means that node i and node j are connected with an undirected edge. G describes a graph among nodes 0, 1, 2, 3, 4, and 5.\\
The edges in G are: (0, 1) (0, 2) (0, 3) (0, 4) (0, 5) (1, 2) (1, 4) (1, 5) (2, 3) (3, 5).\\
\textbf{A}:\\

\bottomrule

\label{table:codegraph_node_count_prompt}
\end{tabular}
\end{table*}

\begin{table*}[!t]
\centering
\small
\caption{An example of \papertitle prompting with the \textit{Adjacency} graph encoding function on the \textit{Edge count} task.} 
\begin{tabular}{p{0.95\linewidth}}
\toprule
\textbf{CodeGraph Prompts for the Edge count Task (1-Shot)}
\\
\midrule
You are tasked with assisting a user who is seeking help with graph networks and Python programming. The user is looking for guidance from an AI that is knowledgeable in graph networks, proficient in Python, and capable of providing bug-free solutions to graph network problems. You will first be given an example to learn about the way to answer. Then you should adhere strictly to the provided program in the example to answer the graph network question.\\
For this task, you will count the number of edges in the undirected graph G.\\
Write a piece of Python code to return the answer in a variable \texttt{`ans'}. Please enclose the code with \texttt{\# CODE START} and \texttt{\# CODE END}. Assume \texttt{edges} is an empty list (\texttt{edges = []}) if not provided.\\
\textbf{Q}: How many edges are in this graph? \\
In an undirected graph, (i,j) means that node i and node j are connected with an undirected edge. G describes a graph among nodes 0, 1, 2, 3, 4, and 5. \\
The edges in G are: (0, 2) (1, 2) (1, 4) (1, 5).\\
\textbf{A}: \begin{verbatim}
# CODE START
from typing import List, Tuple
def count_edges(edges: List[Tuple[str, str]]) -> int:
    unique_edges = set()
    for u, v in edges:
        edge = tuple(sorted((u, v)))
        unique_edges.add(edge)
    return len(unique_edges)
edges = [('0', '2'),
         ('1', '2'),
         ('1', '4'),
         ('1', '5')]
ans = count_edges(edges)
# CODE END
\end{verbatim}
\\
Write a piece of Python code to return the answer in a variable \texttt{`ans'}. Please enclose the code with \texttt{\# CODE START} and \texttt{\# CODE END}. Assume \texttt{edges} is an empty list (\texttt{edges = []}) if not provided.\\
\textbf{Q}: How many edges are in this graph? \\
In an undirected graph, (i,j) means that node i and node j are connected with an undirected edge. G describes a graph among nodes 0, 1, 2, 3, 4, and 5.\\
The edges in G are: (0, 1) (0, 2) (0, 3) (0, 4) (0, 5) (1, 2) (1, 4) (1, 5) (2, 3) (3, 5).\\
\textbf{A}:\\

\bottomrule

\label{table:codegraph_edge_count_prompt}
\end{tabular}
\end{table*}

\begin{table*}[!t]
\centering
\small
\caption{An example of \papertitle prompting with the \textit{Adjacency} graph encoding function on the \textit{Edge existence} task.} 
\begin{tabular}{p{0.95\linewidth}}
\toprule
\textbf{CodeGraph Prompts for the \textit{Edge existence} Task (1-Shot)}
\\
\midrule
You are tasked with assisting a user who is seeking help with graph networks and Python programming. The user is looking for guidance from an AI that is knowledgeable in graph networks, proficient in Python, and capable of providing bug-free solutions to graph network problems. You will first be given an example to learn about the way to answer. Then you should adhere strictly to the provided program in the example to answer the graph network question.\\
For this task, please determine whether there is an an edge between node i and node j in the undirected graph G.\\
Write a piece of Python code to return the answer in a variable \texttt{`ans'}. Please enclose the code with \texttt{\# CODE START} and \texttt{\# CODE END}. Assume \texttt{`edges'} is an empty list (\texttt{edges = []}) if not provided.\\
\textbf{Q}: Is node 8 connected to node 5? \\
In an undirected graph, (i,j) means that node i and node j are connected with an undirected edge. G describes a graph among nodes 0, 1, 2, 3, 4, 5, 6, 7, 8, 9, and 10.\\
The edges in G are: (0, 2) (3, 5) (3, 6) (4, 5) (5, 9).\\
\textbf{A}: \begin{verbatim}
# CODE START
source = '8'
target = '5'
edges = [('0', '2'),
('3', '5'),
('3', '6'),
('4', '5'),
('5', '9')]
def edge_existence(edges, source, target):
    edges_set = set()
    for u, v in edges:
        edges_set.add((u,v))
        edges_set.add((v,u))
    if (source, target) in edges_set: return True
    else: return False
ans = edge_existence(edges,source,target)
# CODE END
\end{verbatim}
\\
Write a piece of Python code to return the answer in a variable \texttt{`ans'}. Please enclose the code with \texttt{\# CODE START} and \texttt{\# CODE END}. Assume \texttt{`edges'} is an empty list (\texttt{edges = []}) if not provided.\\
\textbf{Q}: Is node 3 connected to node 0? \\
In an undirected graph, (i,j) means that node i and node j are connected with an undirected edge. G describes a graph among nodes 0, 1, 2, 3, 4, and 5.\\
The edges in G are: (0, 1) (0, 2) (0, 3) (0, 4) (0, 5) (1, 2) (1, 4) (1, 5) (2, 3) (3, 5).\\
\textbf{A}:\\

\bottomrule

\label{table:codegraph_edge_existence_prompt}
\end{tabular}
\end{table*}

\begin{table*}[!t]
\centering
\small
\caption{An example of \papertitle prompting with the \textit{Adjacency} graph encoding function on the \textit{Node degree} task.} 
\begin{tabular}{p{0.95\linewidth}}
\toprule
\textbf{CodeGraph Prompts for the \textit{Node degree} Task (1-Shot)}
\\
\midrule
You are tasked with assisting a user who is seeking help with graph networks and Python programming. The user is looking for guidance from an AI that is knowledgeable in graph networks, proficient in Python, and capable of providing bug-free solutions to graph network problems. You will first be given an example to learn about the way to answer. Then you should adhere strictly to the provided program in the example to answer the graph network question.\\
For this task, please count the degree of a node in the undirected graph G.\\
Write a piece of Python code to return the answer in a variable \texttt{`ans'}. Please enclose the code with \texttt{\# CODE START} and \texttt{\# CODE END}. Assume \texttt{`edges'} is an empty list (\texttt{edges = []}) if not provided.\\
\textbf{Q}: What is the degree of node 5?  \\
In an undirected graph, (i,j) means that node i and node j are connected with an undirected edge. G describes a graph among nodes 0, 1, 2, 3, 4, and 5.\\
The edges in G are: (0, 1) (0, 5) (2, 5). \\
\textbf{A}: \begin{verbatim}
# CODE START
from collections import defaultdict
from typing import Set, Dict, List, Tuple
def get_adjacency_list( edges: List[Tuple[str, str]]) -> Dict[str, Set[str]]:
    adjacency = defaultdict(set)
    for each_edge in edges:
        u, v = each_edge
        adjacency[u].add(v)
        adjacency[v].add(u)
    return adjacency
def get_node_degree(target_node: str, adjacency_list: Dict[str, Set[str]]) -> int:
    return len(adjacency_list[target_node])
edges = [('0', '1'),
('0', '5'),
('2', '5')]
adjacency_list = get_adjacency_list(edges)
target_node = '5'
ans = get_node_degree(target_node, adjacency_list)
# CODE END
\end{verbatim}
\\
Write a piece of Python code to return the answer in a variable \texttt{`ans'}. Please enclose the code with \texttt{\# CODE START} and \texttt{\# CODE END}. Assume \texttt{`edges'} is an empty list (\texttt{edges = []}) if not provided.\\
\textbf{Q}: What is the degree of node 3?  \\
In an undirected graph, (i,j) means that node i and node j are connected with an undirected edge. G describes a graph among nodes 0, 1, 2, 3, 4, and 5.\\
The edges in G are: (0, 1) (0, 2) (0, 3) (0, 4) (0, 5) (1, 2) (1, 4) (1, 5) (2, 3) (3, 5).\\
\textbf{A}:\\
\bottomrule
\label{table:codegraph_node_degree_prompt}
\end{tabular}
\end{table*}

\begin{table*}[!t]
\centering
\small
\caption{An example of \papertitle prompting with the \textit{Adjacency} graph encoding function on the \textit{Connected nodes} task.} 
\begin{tabular}{p{0.95\linewidth}}
\toprule
\textbf{CodeGraph Prompts for the \textit{Connected nodes} Task (1-Shot)}
\\
\midrule
You are tasked with assisting a user who is seeking help with graph networks and Python programming. The user is looking for guidance from an AI that is knowledgeable in graph networks, proficient in Python, and capable of providing bug-free solutions to graph network problems. You will first be given an example to learn about the way to answer. Then you should adhere strictly to the provided program in the example to answer the graph network question.\\
For this task, please list all nodes that are connected to the specified node in the undirected graph G.\\
Write a piece of Python code to return the answer in a variable \texttt{`ans'}. Please enclose the code with \texttt{\# CODE START} and \texttt{\# CODE END}. Please create an adjacency list from \texttt{`edges'}, default \texttt{`edges'} to an empty list (\texttt{edges = []}) if not provided. \\
\textbf{Q}: List all the nodes connected to 5. \\
In an undirected graph, (i,j) means that node i and node j are connected with an undirected edge. G describes a graph among nodes 0, 1, 2, 3, 4, and 5.\\
The edges in G are: (0, 1) (0, 5) (2, 5). \\
\textbf{A}: \begin{verbatim}
# CODE START
from collections import defaultdict
from typing import Set, Dict, List, Tuple
def get_adjacency_list(edges: List[Tuple[str, str]]) -> Dict[str, 
Set[str]]:
    adjacency = defaultdict(set)
    for each_edge in edges:
        u, v = each_edge
        adjacency[u].add(v)
        adjacency[v].add(u)
    return adjacency
def get_connected_nodes(target_node: str, adjacency_list: Dict[str, Set[str]]) -> str:
    if target_node in adjacency_list and adjacency_list[target_node]:
        connected_nodes = sorted(adjacency_list[target_node], key=lambda x: (x.isdigit(), 
        int(x) if x.isdigit() else x))
        return ', '.join(connected_nodes)
    else:
        return "No nodes"
edges = [('0', '1'),
('0', '5'),
('2', '5')]
adjacency_list = get_adjacency_list(edges)
target_node = '5'
ans = get_connected_nodes(target_node, adjacency_list)
# CODE END
\end{verbatim}
\\
Write a piece of Python code to return the answer in a variable \texttt{`ans'}. Please enclose the code with \texttt{\# CODE START} and \texttt{\# CODE END}. Please create an adjacency list from \texttt{`edges'}, default \texttt{`edges'} to an empty list (\texttt{edges = []}) if not provided. \\
\textbf{Q}: List all the nodes connected to 3.  \\
In an undirected graph, (i,j) means that node i and node j are connected with an undirected edge. G describes a graph among nodes 0, 1, 2, 3, 4, and 5.\\
The edges in G are: (0, 1) (0, 2) (0, 3) (0, 4) (0, 5) (1, 2) (1, 4) (1, 5) (2, 3) (3, 5).\\
\textbf{A}:\\
\bottomrule
\label{table:codegraph_connected_nodes_prompt}
\end{tabular}
\end{table*}

\begin{table*}[!t]
\centering
\small
\caption{An example of \papertitle prompting with the \textit{Adjacency} graph encoding function on the \textit{Cycle check} task.} 
\begin{tabular}{p{0.95\linewidth}}
\toprule
\textbf{CodeGraph Prompts for the \textit{Cycle check} Task (1-Shot)}
\\
\midrule
You are tasked with assisting a user who is seeking help with graph networks and Python programming. The user is looking for guidance from an AI that is knowledgeable in graph networks, proficient in Python, and capable of providing bug-free solutions to graph network problems. You will first be given an example to learn about the way to answer. Then you should adhere strictly to the provided program in the example to answer the graph network question.\\
For this task, please determine if the given undirected graph G contains any cycles.\\
Write a piece of Python code to return the answer in a variable \texttt{`ans'}. Please enclose the code with \texttt{\# CODE START} and \texttt{\# CODE END}. Assume \texttt{`edges'} and \texttt{`nodes'} are empty lists (\texttt{edges = [], nodes=[]}) if not provided.\\
\textbf{Q}: Is there a cycle in this graph? \\
In an undirected graph, (i,j) means that node i and node j are connected with an undirected edge. G describes a graph among nodes 0, 1, 2, 3, 4, and 5.\\
The edges in G are: (0, 2) (1, 2) (1, 4) (1, 5). \\
\textbf{A}: \begin{verbatim}
# CODE START
def has_cycle(graph):
    visited = set()
    for node in graph:
        if node not in visited:
            if dfs(graph, node, visited, None):
                return True
    return False
def dfs(graph, node, visited, parent):
    visited.add(node)
    for neighbor in graph[node]:
        if neighbor not in visited:
            if dfs(graph, neighbor, visited, node):
                return True
        elif neighbor != parent:
            return True
    return False
nodes = ['0','1','2','3','4','5']
edges = [('0', '2'),
('1', '2'),
('1', '4'),
('1', '5')]
graph = {node: [] for node in nodes}
for edge in edges:
     graph[edge[0]].append(edge[1])
if has_cycle(graph):
    ans = "Has cycle."
else:
    ans = "No cycle."
# CODE END
\end{verbatim}
\\
Write a piece of Python code to return the answer in a variable \texttt{`ans'}. Please enclose the code with \texttt{\# CODE START} and \texttt{\# CODE END}. Assume \texttt{`edges'} and \texttt{`nodes'} are empty lists (\texttt{edges = [], nodes=[]}) if not provided.\\
\textbf{Q}: Is there a cycle in this graph? \\
In an undirected graph, (i,j) means that node i and node j are connected with an undirected edge. G describes a graph among nodes 0, 1, 2, 3, 4, and 5.\\
The edges in G are: (0, 1) (0, 2) (0, 3) (0, 4) (0, 5) (1, 2) (1, 4) (1, 5) (2, 3) (3, 5).\\
\textbf{A}:\\
\bottomrule
\label{table:codegraph_cycle_check_prompt}
\end{tabular}
\end{table*}

\section{Dataset Statistics}
The graphs used in this paper are taken from the GraphQA benchmark~\cite{fatemi2023talk}. We evaluate the average number of nodes, the average number of edges, and the average node degree for different graphs, which are generated by various graph generators and used in the experiments of this paper, as shown in Table~\ref{table:dataset_statistics}.

\begin{table*}[t]
\small
\centering
\footnotesize
\resizebox{\textwidth}{!}{\setlength{\tabcolsep}{4pt}

\begin{tabular}{ll|ccc|ccc} 
\toprule
 & & \multicolumn{3}{c|}{\textbf{Train Set}} & \multicolumn{3}{c}{\textbf{Test Set}} \\
\cmidrule(lr){3-5} \cmidrule(lr){6-8}
& \textbf{Generator} & \textbf{Avg. Nodes} & \textbf{Avg. Edges} & \textbf{Avg. Degree} & \textbf{Avg. Nodes} & \textbf{Avg. Edges} & \textbf{Avg. Degree} \\
\midrule
\scriptsize{\multirow{7}{*}{}} 
& ER & 11.78 & 38.07 & 5.54 & 12.37 & 39.79 & 5.70 \\
& BA & 11.77 & 28.30 & 4.29 & 12.25 & 30.02 & 4.40 \\
& SBM & 11.69 & 34.91 & 5.19 & 12.27 & 38.68 & 5.59 \\
& Star & 11.81 & 10.81 & 1.80 & 12.31 & 11.31 & 1.81 \\
& SFN & 11.81 & 15.11 & 2.51 & 12.31 & 15.94 & 2.55 \\
& Path & 11.81 & 10.81 & 1.80 & 12.31 & 11.31 & 1.81 \\
& Complete & 11.81 & 72.95 & 10.81 & 12.31 & 78.54 & 11.31 \\
\bottomrule
\end{tabular}
}

\caption{Summarization of dataset statistics for various graph structures in the GraphQA dataset.}
\label{table:dataset_statistics}

\end{table*}

\end{document}